\documentclass{article}

\usepackage[preprint]{neurips_2026}
\makeatletter
\renewcommand{\@notice}{}
\makeatother

\usepackage[utf8]{inputenc}
\usepackage[T1]{fontenc}
\usepackage{hyperref}
\usepackage{url}
\usepackage{booktabs}
\usepackage{microtype}
\usepackage{xcolor}

\usepackage{amsmath,amssymb}
\usepackage{graphicx}
\usepackage{caption}
\usepackage{makecell}
\usepackage{array}
\newcolumntype{L}[1]{>{\raggedright\arraybackslash}p{#1}}
\usepackage{enumitem}
\usepackage{xspace}
\usepackage{pifont}
\usepackage[capitalize,noabbrev]{cleveref}
\usepackage{algorithm}
\usepackage[noend]{algpseudocode}

\definecolor{linkblue}{rgb}{0.05,0.25,0.55}
\hypersetup{colorlinks=true, linkcolor=linkblue, citecolor=linkblue, urlcolor=linkblue,
  pdftitle={Build, Don't Play: Self-Improving LLM Research Agents Develop a Hearthstone Battlegrounds
  AI That Rivals Top Players}}

\makeatletter\g@addto@macro{\UrlBreaks}{\UrlOrds}\makeatother

\graphicspath{{figures/}}

\newcommand{\frameworkname}{\textsc{AutoTavern}\xspace}
\newcommand{\envname}{\textsc{TavernGym}\xspace}
\newcommand{\agentname}{\textsc{TavernAgent}\xspace}
\newcommand{\benchname}{\textsc{TavernBench}\xspace}
\newcommand{\game}{Hearthstone Battlegrounds\xspace}
\newcommand{\cmark}{\ding{51}}
\newcommand{\xmark}{\ding{55}}

\title{\fontsize{16}{19.5}\selectfont Build, Don't Play:\\
Self-Improving LLM Research Agents Develop a\\
Hearthstone Battlegrounds AI That Rivals Top Players}

\author{%
  Changlian Ma \\
  Nanjing University \\
  Shanghai AI Laboratory \\
  \texttt{changlianma@smail.nju.edu.cn}
  \And
  Limin Wang \\
  Nanjing University \\
  Shanghai AI Laboratory \\
  \texttt{lmwang@nju.edu.cn}
}

\begin{document}

\maketitle

\begin{abstract}
  Can LLM agents carry a hard real-world problem from raw data to a system that holds its own
  against people? We study this in \game, an eight-player auto-battler with stochastic shops and
  battles, hidden information and nearly a thousand interacting cards. We present
  \frameworkname, a self-improving research system: specialised LLM research agents share a typed
  research memory, every claim passes verification gates, and live games close the loop, so the agents improve the simulator, opponents and policy and, from the same feedback,
  their own research process; a human sets goals and rules on ambiguous mechanics. In 20 days \frameworkname built \envname, the
  first faithful, full-content reproduction of the live game, and delivered 23 versions of
  \agentname, trained with PPO against opponents calibrated to real games.
  \agentname plays complete games on the live ladder by itself: its rating rose from 6,000 to about 15,000 and, on the public leaderboard's measure, its last 20 records average a
  placement of 1.60 with 80\% first places, better than the server's ten highest-rated players on
  every statistic (2.34, 59\%). \benchname, 960 tasks in 40
  scenarios, shows that the system the agents built is the strongest player: \agentname answers
  91\% of the play tasks against at most 63\% for frontier LLMs, and decides in
  0.03~s where they need up to 26~s; within the second a live turn allows, it keeps 90\% and no
  LLM more than 26\%. LLM agents build far better than they play. We will open-source \benchname
  and \envname at \mbox{\url{https://github.com/MCG-NJU/AutoTavern}}.
\end{abstract}

\section{Introduction}
\label{sec:intro}

\begin{figure}[t]
  \centering
  \includegraphics[width=\linewidth]{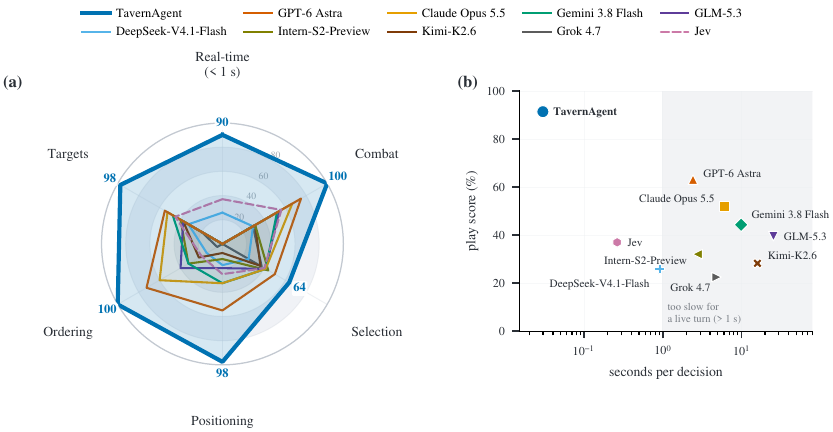}
  \caption{\textbf{\benchname at a glance.} (a)~Accuracy (\%) per capability, with the play score
  counting only answers within the second a live turn allows per decision; numbers: \agentname. (b)~Play score against seconds per
  decision; shaded: too slow for a live turn. All systems are in \cref{tab:bench-results}.}
  \label{fig:teaser}
\end{figure}

Games have long served as milestones for artificial intelligence, from Go and chess
\citep{silver2016mastering,silver2018general} to StarCraft~II, Dota~2, poker, Mahjong and
Stratego \citep{vinyals2019grandmaster,berner2019dota,brown2019superhuman,li2020suphx,perolat2022mastering}.
Each of these systems was the product of a large team of human experts who built the
environment, designed the learning algorithm and debugged the countless ways in which a
simulated training world differs from the real game. Today, LLM agents write and run much of
the code behind such systems, and a growing line of work asks whether they can automate the
research itself \citep{lu2024aiscientist,yamada2026towards,wijk2025rebench,srikanth2026recursive}.
Existing evidence, however, comes from benchmark tasks or paper reviews; it remains open whether
agents can carry a hard, real-world problem from raw data to a system that holds its own against
people, where performance is ultimately measured in the real game.

We study this question in \game, Blizzard's eight-player auto-battler, played by millions. Each
turn a player spends gold in a randomly stocked tavern to buy, sell and position minions, then
fights an automatic battle against one opponent; players are eliminated in turn until one
remains. The game combines stochastic shops and battles, hidden information about seven
opponents, a shared card pool, a horizon of hundreds of decisions with a reward only at the end,
and a content base of nearly one thousand interacting cards, heroes and trinkets that changes
with every patch. No faithful simulator of the live game has been available, and existing AI
work either advises human players or studies simplified versions of the game
(\cref{sec:related}).

\begin{figure}[t]
  \centering
  \includegraphics[width=\linewidth]{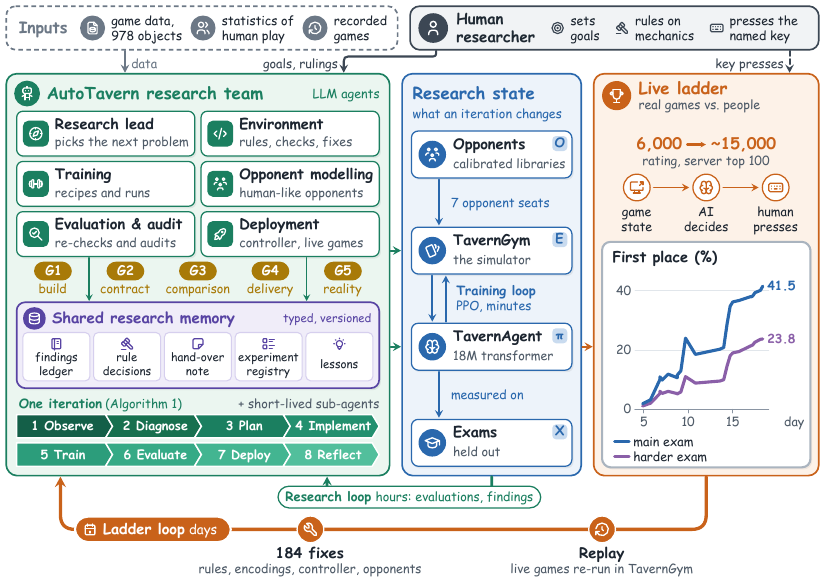}
  \caption{\textbf{Overview of \frameworkname, a self-improving research system.} Six LLM research
  agents share one research memory and run three loops: training (minutes), research (hours) and the live ladder (days). The human sets
  goals, rules on ambiguous mechanics and presses the keys the AI names. Chart: first-place rate of
  the 23 delivered \agentname versions on one simulator build and three unseen seeds (24,576 games
  per point).}
  \label{fig:overview}
\end{figure}

We present \frameworkname, a self-improving research system in which a team of specialised LLM
research agents iterates a model autonomously (\cref{fig:overview}). Its principle is
reality-anchored self-improvement. Agents for the environment, training, opponent modelling,
evaluation and deployment communicate only through a typed, versioned research memory; five
verification gates stand between every claim and that memory; and live games on the real ladder
close the loop, turning discrepancies between simulation and reality into fixes of rules, input
encodings and opponent models. The same feedback improves the agents' research process: errors
they catch become lessons in the memory, loaded by every later session, or new gates, so each
iteration starts from a better world and a better method. Because the loop does not pay for the
same mistake twice, it took the agent from its first version to the level of the server's best
players in 20 days. The human sets goals and rules on ambiguous mechanics. Over 20 days the
framework produced:

\begin{itemize}[leftmargin=*,itemsep=1pt,topsep=2pt]
  \item \textbf{\frameworkname}, a self-improving research system in which LLM research agents
  improve a model, its training world and their own research process from the feedback of a real
  environment. In 20 days it ran about 10,000 experiments and delivered 23 agent versions with 418
  human messages in all, and the errors it caught became lessons and gates that changed how it
  works (\cref{sec:framework,sec:lessons}).
  \item \textbf{\envname}, the first faithful, full-content reproduction of the
  live game: all 978 game objects of patch 36.6.3, deterministic, 190--280 complete
  eight-player games per second on a single core, and aligned rule by rule with real games
  (\cref{sec:env}). We will open-source it.
  \item \textbf{\agentname}, the first Battlegrounds AI to play complete games
  on the live ladder, and the strongest. Its 23 successive versions, each passing an evaluation gate, raised the
  first-place rate against reality-calibrated opponents from 1.9\% to 41.5\% (random legal play
  and a scripted bot: 0\%), and on the live ladder the deployed agent climbed by itself from a rating of
  6,000 to about 15,000; on the public leaderboard's measure its recent results are better than
  the server's ten highest-rated players' on every statistic (\cref{sec:agent} and \cref{app:deploy}).

  \item \textbf{\benchname}, the first decision benchmark for Battlegrounds: 960 tasks in
  40 scenarios across eight categories, labelled by the simulator
  (\cref{sec:bench}). We will open-source it.
\end{itemize}

The benchmark bears out our title (\cref{fig:teaser}): asked to make the decisions
themselves, the LLMs that built a ladder-level agent fall well short of the system they built, and
decide far too slowly for a live turn (\cref{sec:bench}). LLMs build far better than they play.

\section{Related Work}
\label{sec:related}

\paragraph{AI for Hearthstone and Battlegrounds.}
Research on Hearthstone has focused on the two-player constructed mode
\citep{dockhorn2019introducing}. For Battlegrounds, \citet{zolboot2022hearthstone} built a
Monte Carlo tree search assistant on a simplified simulator, evaluated on the first turns of
recorded games; an open-source simulator for reinforcement learning omits heroes
\citep{bgsimulator}; community tools such as Firestone and Bob's Buddy simulate
only the combat phase to show win odds \citep{firestone,bobsbuddy}; and DeepBattler uses an
LLM to give players advice during live games without an environment or an agent of its own
\citep{deepbattler}. None reproduces the full live game or trains an agent that plays it
(\cref{tab:envs}).

\paragraph{Large-scale game RL.}
Superhuman systems for Go, StarCraft~II, Dota~2, Honor of Kings, poker, Mahjong, DouDizhu and
Stratego \citep{silver2016mastering,vinyals2019grandmaster,berner2019dota,ye2020towards,brown2019superhuman,li2020suphx,zha2021douzero,perolat2022mastering}
combine self-play or human data with large-scale training and start from an existing
environment. We use a standard learner, PPO \citep{schulman2017proximal} with a transformer
\citep{vaswani2017attention}, and focus on building and correcting the training world itself,
which here is done by agents.

\paragraph{LLM research agents.}
The AI Scientist automates idea generation, experiments and paper writing
\citep{lu2024aiscientist,yamada2025aiscientistv2,yamada2026towards}; Eureka and AlphaEvolve use
LLMs to propose reward functions and programs that an automatic evaluator scores
\citep{ma2024eureka,novikov2025alphaevolve}; MLAgentBench, MLE-bench, RE-Bench and PaperBench
measure agents on machine-learning tasks
\citep{huang2024mlagentbench,chan2025mlebench,wijk2025rebench,starace2025paperbench}; and recent
systems orchestrate coding agents or improve their own scaffolds
\citep{lange2026agentic,zhang2025dgm,srikanth2026recursive}; \citet{kirgis2026can} find that
agents complete the engineering of open-ended projects but struggle with the research questions.
\frameworkname differs in what improves and in how the improvement is judged. Its agents improve
their research process through a memory of lessons and new verification gates rather than by
rewriting their own code. Progress is measured on a live ladder against people, a reality loop
re-anchors simulator and opponents to live games against proxy drift, and verification gates stand
between every claim and the shared memory against false findings.

\paragraph{LLMs as game players.}
Benchmarks evaluate LLMs on text, puzzle, video and strategic games
\citep{wu2024smartplay,duan2024gtbench,costarelli2024gamebench,paglieri2025balrog,guertler2025textarena,hu2025lmgame}
and on poker decisions \citep{zhuang2025pokerbench}. \benchname adds a stochastic multiplayer
game of economy and combat with answers verified by a simulator, a strong specialised agent as reference
and a real-time view: the decisions a model gets right within a live turn's time.

\section{Hearthstone Battlegrounds}
\label{sec:game}

A game has eight players, each with a hero, 30--40 health and a Hero Power. Every turn consists of
a \emph{recruit} phase and a \emph{combat} phase. In the recruit phase the player spends gold
(3 on turn 1, rising to 10) to buy minions from a tavern of random offers drawn from a pool
shared by all players, sell minions, refresh or freeze the offers, upgrade the tavern tier
(1--7, unlocking stronger minions), play cards from the hand, use the Hero Power and arrange
the warband of up to seven minions. Three copies of a minion merge into a golden version that
offers a discovery from the next tier. Battlecries, deathrattles, auras, magnetic attachments,
tavern spells, trinkets and other mechanics interact across turns. In combat, the warband fights
one opponent automatically, with random targets; the loser takes damage, and players are
eliminated until one remains. The outcome is the placement, 1 to 8.

We model a game as a partially observable stochastic game with a terminal utility of the
placement. A player observes its own state and, for each opponent, only the public leaderboard
and the warband last seen in combat. The current patch (36.6.3) contains 978 game objects: minions, heroes, trinkets, spells and
other cards. A game lasts about 15 turns and about 450 atomic actions per player.

\section{The \frameworkname System}
\label{sec:framework}

\frameworkname is a self-improving research system: LLM research agents improve a model against an
environment they cannot simulate exactly and, from the same feedback, their own research process.
Its central idea is \emph{reality-anchored self-improvement}: every proxy the agents optimise (the
simulator, the opponent model, the exams) is re-anchored to the real game at a fixed cadence, every
claim must survive a check designed to refute it before it changes the shared state, and the
errors these checks catch change the process that made them.

\paragraph{Problem setting.} The research state at iteration $t$ is a set of artifacts: an
environment $E_t$, an opponent model $O_t$, a set of exams $X_t$ and a policy $\pi_t$, together
with the research process itself: the memory $M_t$ of findings and lessons and the gates $G_t$. The goal is
the policy's real performance $J(\pi)$, its expected placement on the live ladder, observable only
through a few slow, costly games. Inside the loop the agents can only improve a
proxy $\hat J(\pi; E_t, O_t, X_t)$. Two failure modes follow. \emph{Proxy drift}: $\hat J$ improves
while $J$ does not, because $E_t$ or $O_t$ is wrong in a way the policy exploits. \emph{False
findings}: a reading that reflects noise, a leak or selection is recorded as a result and steers
later work. \frameworkname answers the first with a reality loop and the second with
verification gates; both act on a shared memory, the only channel through which findings pass between agents.
Self-improvement closes the loop: an iteration updates the artifacts and, when it catches an error,
$M_t$ and $G_t$, so that later iterations do not repeat it.

\paragraph{Role agents.} Long-running LLM research agents own the research roles (\cref{fig:overview},
\cref{tab:roles}). A \emph{research lead} chooses the next problem and dispatches work. An
\emph{environment agent} implements and verifies rules and repairs discrepancies found in live
games, and an \emph{opponent agent} builds human-like opponents from human statistics and live
games. A \emph{training agent} designs recipes and curricula, an \emph{evaluation agent}
measures, re-checks and audits, and a \emph{deployment agent} maintains the client controller and
the models' input formats. Short-lived workers run probes and implement features.

\paragraph{Typed research memory.} Agents never talk directly; they read and write versioned,
typed stores: a ledger of numbered findings in which a correction is a new entry pointing back, a
registry of rule decisions and open questions, a hand-over note, an experiment registry, and a
store of lessons loaded at the start of every session. A finding is never edited away, so the
memory records how conclusions changed as well as what they are.

\paragraph{Verification gates.} Five gates stand between an idea and the shared state:
\begin{itemize}[leftmargin=*,itemsep=0pt,topsep=1pt]
  \item \textsc{G1}~build: a code change passes the full test suite before it reaches the shared build;
  \item \textsc{G2}~contract: every model is tied to the simulator build it was trained on, and the
  controller refuses a model trained for another;
  \item \textsc{G3}~comparison: models are compared only within one build, over several seeds;
  \item \textsc{G4}~delivery: a candidate is delivered only after a re-check on seeds it was not selected on;
  \item \textsc{G5}~reality: delivered models play the live ladder, and their games are replayed in the simulator.
\end{itemize}
The gates acted throughout: candidates were held back after re-checks, findings were revised by
later ones, and live games led to 184 fixes of rules, input encodings, the client controller and
the opponents, counted in \cref{app:framework}.

\paragraph{Reality loop.} The controller logs every live game; game records are linked to the
agent's decisions; the evaluation agent replays real combats and real recruit steps in \envname and
reports where simulation and reality part; the responsible agent fixes rules, input encodings or
opponents, and the next iteration trains on the corrected world (\cref{sec:lessons}).

\paragraph{Self-improvement.} The loop improves the research process as well as its products. A
finding that overturns an earlier one is recorded as a correction pointing back. A mistake the
agents trace to their own method becomes a lesson that every later session loads at its start, or
a new gate. Of the six errors in \cref{sec:lessons}, two became gates (the re-check on unseen
seeds, \textsc{G4}, and the comparison of the two input encoders in \textsc{G2}) and two became
standing rules (reseed copied games, and calibrate opponents on those the learner meets). Because a
caught error changes the process and not only the current result, the loop does not pay for the
same mistake twice: over the 20 days the 23 delivered versions raised the first-place rate from
1.9\% to 41.5\% (\cref{fig:overview}).

\subsection{The iteration cycle}

\cref{alg:loop} shows one iteration. The cycle runs at three time scales: policy updates every
few seconds to minutes (the training loop), experiments and findings within hours (the research
loop), and live games feeding back into the environment within days (the ladder loop). A
scheduler runs the registered experiments within a resource budget and keeps their successors
queued, so the loop runs for days without a human present. The human researcher enters at two
points only: setting goals, and ruling on mechanics that the game data leave ambiguous.

\begin{algorithm}[t]
\caption{One iteration of \frameworkname}
\label{alg:loop}
\small
\begin{algorithmic}[1]
\State \textbf{Observe:} read the research memory, the latest evaluations and the latest live games
\State \textbf{Diagnose:} attribute the largest gap to rules, input encoding, opponent model, policy or goal
\State \textbf{Plan:} choose an intervention (fix the environment, recalibrate opponents, change the recipe, add a label)
\State \textbf{Implement:} the responsible agent builds it in isolation and passes the test gate
\State \textbf{Train:} register runs; the scheduler executes them and queues successors
\State \textbf{Evaluate:} compare within one build; re-check on unseen seeds; deliver only if better
\State \textbf{Deploy:} the delivered model plays on the live ladder; game records flow back
\State \textbf{Reflect:} record findings, correct earlier conclusions, turn caught errors into lessons and gates
\end{algorithmic}
\end{algorithm}

\subsection{Autonomy in numbers}

\cref{tab:autonomy} summarises the 20 days. Agents registered about 10,000 experiments, and 23 agent
versions passed the delivery gate. The human typed 418 messages over the whole project, 0.6\% of
all messages. Labelled by their main intent (automatically, checked by hand on a sample), 42\% ask
for progress or a report, 24\% set or change a goal, 11\% concern resources, 8\% approve a proposal,
6\% correct an agent, 5\% rule on a game mechanic and 5\% report live games.

\begin{table}[t]
  \centering
  \caption{\textbf{Autonomous research in numbers.}}
  \label{tab:autonomy}
  \small
  \begin{tabular}{@{}lr@{\hspace{1em}}lr@{}}
    \toprule
    Duration (from scratch to ladder) & 20 days & Experiments registered & about 10,000 \\
    Engine code (Rust) & 63k lines & Delivered agent versions & 23 \\
    Rule verification records & 375 & Fixes from live games & 184 \\
    Human messages & 418 (0.6\% of all messages) & & \\
    \bottomrule
  \end{tabular}
\end{table}

\section{\envname: Reproducing the Real Game}
\label{sec:env}

\paragraph{Design.} \envname is a Rust engine with thin Python bindings. One effect context
serves the recruit and the combat phase, so every card is a table of event hooks that work the
same in both; the state is an arena of entities, cheap to copy and serialisable; all randomness
comes from one seeded generator, so a seed and an action sequence reproduce a game bit for bit.
Card data are generated from the game's own files, later patches are overlays, and a rule hash
identifies every build.

\paragraph{Fidelity to the live game.} All 978 objects of patch 36.6.3 are implemented. 375
verification records and 220 confirmed rule decisions document how each mechanic was checked, and
unit tests, fuzzing with invariant checks after every action and an exact legal-action mask guard
the implementation. Live games led to 46 fixes of rules and card data and 73 of input encodings,
found by replaying live turns, combats and offers in the engine (\cref{sec:lessons,app:framework}).

\paragraph{Speed and interface.} A complete eight-player game with scripted bots takes 3.6--5.3~ms
on one core (190--280 games per second), and a late-game combat 13--19~\textmu s (52--76
thousand per second). The RL interface exposes 245 discrete actions with an exact legal mask, a
player-view observation with only the information visible in the real client, batched
environments and exact save and restore of running games for lookahead and replays.

\begin{table}[t]
  \centering
  \caption{\textbf{Battlegrounds environments and agents.} Only \envname reproduces the full live
  game, and only \agentname plays real games.}
  \label{tab:envs}
  \small
  \resizebox{\linewidth}{!}{%
  \begin{tabular}{@{}lcccccc@{}}
    \toprule
    & \makecell{Full content,\\current patch} & Heroes & \makecell{Recruit +\\combat} &
      \makecell{Trainable\\batch API} & \makecell{Aligned with\\live games} & \makecell{Plays live\\games} \\
    \midrule
    MCTS assistant \citep{zolboot2022hearthstone} & \xmark & partial & \cmark & \xmark & \xmark & \xmark \\
    BGSimulator \citep{bgsimulator} & \xmark & \xmark & \cmark & \cmark & \xmark & \xmark \\
    Firestone / Bob's Buddy \citep{firestone,bobsbuddy} & \cmark & \cmark & \xmark & \xmark & \cmark & \xmark \\
    DeepBattler \citep{deepbattler} & n/a & n/a & n/a & \xmark & n/a & \xmark \\
    \midrule
    \envname + \agentname (ours) & \cmark & \cmark & \cmark & \cmark & \cmark & \cmark \\
    \bottomrule
  \end{tabular}}
\end{table}

\section{\agentname: Learning to Play the Live Game}
\label{sec:agent}

\paragraph{Model and training.} \agentname is an entity transformer: each own minion, hand card,
tavern offer, pending choice and opponent is a token (up to 56), a card is represented by an
identity embedding, static features and an embedding of its rule text, and a pointer head
produces the 245 masked action logits next to value and placement heads (d=384, 8 layers, 8
heads, 18M parameters, the best of three sizes in an early comparison, \cref{tab:model-size}).
It is trained with PPO \citep{schulman2017proximal} on about one million
decisions per iteration with a terminal placement utility; each
training stage of 300 iterations covers about 310 million decisions (570 thousand games) and
starts from the previous stage under a KL anchor.

\paragraph{Reality-calibrated opponents.} Rather than self-play, the learner faces seven
\emph{library seats}: each turn, a seat takes a board (with hand and hidden counters) from a library
built from statistics of top human players and recalibrated, turn by turn, to the opponents the
agent actually met in live games; exam libraries are never used for training. This was the largest
single gain: from the same checkpoint and with the same budget, training against the
calibrated libraries raises the first-place rate from 20.9\% to 27.2\% on the main exam and from
10.3\% to 13.5\% on the harder one (\cref{tab:agent}).

\paragraph{Iterative improvement.} The chart in \cref{fig:overview} traces all 23 delivered versions (TA-1 to
TA-23, \cref{tab:releases}), re-measured on one simulator build, one exam and three unseen seeds:
the first-place rate rises from 1.9\% to 41.5\% and the average placement from 3.40 to 2.33
(\cref{tab:agent}).

\begin{table}[t]
  \centering
  \caption{\textbf{\agentname against baselines, and the opponent ablation.} One simulator build,
  three unseen seeds, 24,576 games per row, seven reality-calibrated opponent seats; main exam
  (left) and the harder exam (right). The ablation pair shares its starting point and training
  budget.}
  \label{tab:agent}
  \footnotesize
  \setlength{\tabcolsep}{4pt}
  \begin{tabular}{@{}lcccccc@{}}
    \toprule
    & \multicolumn{3}{c}{Main exam} & \multicolumn{3}{c}{Harder exam} \\
    \cmidrule(lr){2-4}\cmidrule(l){5-7}
    Agent & Place $\downarrow$ & First \% & Top-4 \% & Place $\downarrow$ & First \% & Top-4 \% \\
    \midrule
    Random legal actions & 7.97 & 0.0 & 0.0 & 7.97 & 0.0 & 0.0 \\
    Scripted bot & 7.54 & 0.0 & 0.5 & 7.61 & 0.0 & 0.2 \\
    \agentname, first version (TA-1) & 3.40 & 1.9 & 83.4 & 4.07 & 1.1 & 65.4 \\
    \agentname, mid version (TA-11) & 2.83 & 18.5 & 85.5 & 3.75 & 8.9 & 67.8 \\
    \agentname, final (TA-23) & \textbf{2.33} & \textbf{41.5} & \textbf{89.0} & \textbf{3.28} & \textbf{23.8} & \textbf{73.2} \\
    \midrule
    \multicolumn{7}{@{}l}{\emph{Opponents} (from the same checkpoint, 150 iterations each)} \\
    \quad uncalibrated opponent libraries & 2.74 & 20.9 & 86.8 & 3.64 & 10.3 & 69.6 \\
    \quad reality-calibrated libraries & \textbf{2.65} & \textbf{27.2} & 86.6 & \textbf{3.59} & \textbf{13.5} & 69.2 \\
    \bottomrule
  \end{tabular}
\end{table}

\begin{figure}[t]
  \centering
  \includegraphics[width=\linewidth]{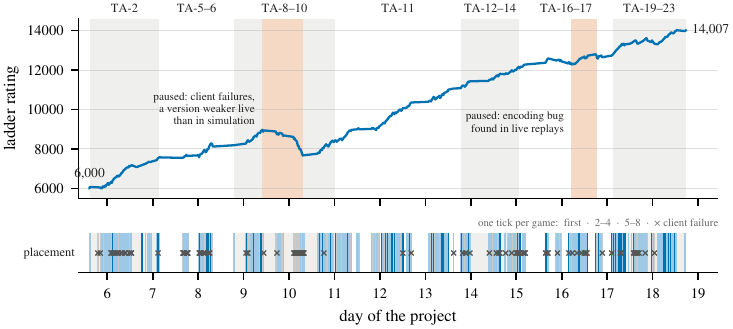}
  \caption{\textbf{The climb on the live ladder.} Rating after each of 392 games, days 5--18.
  Grey bands: the versions in play (\cref{tab:ladder}); orange: the two pauses in which the loop
  fixed what the ladder had exposed (\cref{sec:lessons}). Strip: one tick per game with its
  placement; $\times$: client failure.}
  \label{fig:ladder}
\end{figure}

\paragraph{Live ladder.} The deployed agent plays the live ranked ladder by itself: the
controller feeds the live game's state to the policy in the environment's own encoding, the policy
decides every operation, and a human presses the shortcut key the AI names, a feature of the game's
official plug-in (\cref{app:deploy}). Most delivered versions entered this loop, so the agent's
iterations are also its climb on the ladder (\cref{fig:ladder,tab:ladder}): over 392 games on
days 5--18 of the project the rating rose from 6,000 to 14,007. Games in which the client failed (96, flagged by
the controller's logs, never by the result; \cref{app:deploy}) measure the client rather than the
agent; without them the average placement is 3.22, with 16.4\% first places and 77.1\% top-four finishes. The latest versions
(TA-19--23) averaged 2.37 with 32.6\% first places and 91.3\% top-four. A problem the ladder exposed
paused testing until the loop fixed it (\cref{fig:ladder}). By the end of the study the agent was rated about
15,000, in the server's top 100; on the public leaderboard's measure its last 20 records average
1.60 against 2.34 for the ten highest-rated players, better on every statistic, with no sign of
a plateau (\cref{tab:top10}).

\section{\benchname: Build, Don't Play}
\label{sec:bench}

\paragraph{Design.} \benchname asks whether general-purpose LLMs can make the decisions that
\agentname makes: 960 tasks in 40 scenarios and eight categories (\cref{tab:bench-cats}). A task
gives the player-visible state as text, the rule text of every card shown, an objective and an
answer format. Labels come from \envname: rule tasks execute the actions; combat tasks simulate
4,096 combats; short-term decisions are scored by the simulated win rate of the next combat after
the chosen option (correct within three points of the best); strategic decisions are turn
plans cut from natural games (upgrade, buy, refresh or save the Gold), each played out 128 times
by an earlier version (TA-21) and kept when one is better by a quarter place and two standard
errors; hero and trinket choices are scored against
top human players (\cref{app:bench}).

\paragraph{\agentname's interface.} \agentname does not read the text: it receives the task's
engine state as in a game. Its policy answers the choices judged by playing the game out, its value
head the hero offers. Combat questions and short-term decisions use \envname as the agent's
model of the game: it applies each option and simulates 256 combats after it (one-step
lookahead). Rule tasks ask what the environment does, which \envname answers by construction;
\agentname is not scored on them, and its row shows a dash.

\paragraph{Protocol.} LLMs answer with one JSON object within a fixed output budget, reasoning
included (more is scored as wrong), in the fastest mode that answered a pilot of one task from each
scenario (\cref{app:bench}).
All systems answer the same 368 tasks (eight per scenario, all play-out tasks). Jev
\citep{typesafejev}, a decision model that picks among typed options without generating text, gets
the same text with each task posed as a choice over its candidates (\cref{app:bench}).

\begin{table}[t]
  \centering
  \caption{\textbf{\benchname results} (368 tasks; accuracy in \%). Play: the 30 scenarios
  \agentname answers, mean over scenarios.
  Real-time: the play score counting only answers given within 1~s. Rules: the ten rule-execution
  scenarios (answered by \envname by construction, so not scored for \agentname). Time: median
  seconds per task (LLMs: estimated from Artificial Analysis medians, \cref{app:bench}; others
  measured).}
  \label{tab:bench-results}
  \footnotesize
  \setlength{\tabcolsep}{2.6pt}
  \begin{tabular}{@{}lrrrrrrrrrr@{}}
    \toprule
    System & Play & Real-time & Combat & Position & Order & Targets & Select & Strategy & Rules & Time (s) \\
    \midrule
    \agentname{} & 91 & 90 & 100 & 98 & 100 & 98 & 64 & 75 & -- & 0.03 \\
    \midrule
    Claude Opus 5.5 & 52 & 0 & 67 & 32 & 60 & 52 & 41 & 38 & 92 & 6.2 \\
    Claude Sonnet 5.5 & 53 & 0 & 58 & 45 & 68 & 52 & 39 & 29 & 85 & 1.4 \\
    GPT-6 Astra & 63 & 0 & 75 & 55 & 72 & 55 & 50 & 54 & 95 & 2.4 \\
    GPT-6.1 Sol & 46 & 0 & 50 & 42 & 48 & 28 & 55 & 43 & 56 & 2.0 \\
    Gemini 3.8 Flash & 44 & 0 & 52 & 32 & 32 & 48 & 41 & 43 & 90 & 9.9 \\
    Grok 4.7 & 22 & 0 & 29 & 0 & 5 & 20 & 38 & 37 & 79 & 4.8 \\
    GLM-5.3 & 40 & 0 & 54 & 20 & 40 & 38 & 41 & 34 & 65 & 26 \\
    Kimi-K2.6 & 28 & 0 & 31 & 8 & 22 & 38 & 36 & 27 & 56 & 16 \\
    Intern-S2-Preview & 32 & 0 & 31 & 12 & 32 & 40 & 44 & 34 & 54 & 2.8 \\
    DeepSeek-V4.1-Flash & 26 & 26 & 29 & 18 & 15 & 32 & 25 & 19 & 29 & 0.91 \\
    Jev & 37 & 37 & 56 & 25 & 20 & 45 & 40 & 17 & 35 & 0.26 \\
    \bottomrule
  \end{tabular}
\end{table}

\paragraph{Results.} \Cref{tab:bench-results} and \cref{fig:teaser} give four findings.
\emph{The system the LLM agents built is the strongest player.} \agentname answers 91\% of the play tasks;
the best LLM playing directly, GPT-6 Astra, 63\%, the Claude models 52--53\%, GPT-6.1 Sol 46\%,
Gemini~3.8 Flash 44\% and the remaining models 22--40\%. It is near-perfect on combat, positioning,
ordering and targets, where LLMs must simulate fights in their head; with \envname as a tool they
recover most of that gap (GPT-6 Astra 62\% to 97\%, \cref{tab:tools}).

\emph{Latency rules out live play for the LLMs.} A late recruit turn lasts about two minutes and holds up to about a hundred
operations, so a decision that takes more than a second cannot be played. \agentname answers a task in 0.03~s
(\cref{tab:bench-results}). Gateway timings are not
comparable across providers, so we time every LLM on one standard footing \citep{artificialanalysis}:
the median time to its first answer token, which includes the model's typical thinking, plus the
task's answer at its median output speed. Their median task takes 1.4--26~s, DeepSeek-V4.1-Flash
without reasoning 0.9~s. Within the one-second budget \agentname keeps 90\%,
DeepSeek its 26\% and Jev its 37\%, and every other LLM falls to 0\%. Speed alone would not suffice: a turn chains dozens of decisions into one plan, so a model that errs on a third of
them does not play a third worse, it loses the plan; per-decision errors compound over a turn
rather than average out.

\emph{Reading rules is not playing.}
The frontier LLMs execute rules well (GPT-6 Astra 95\%, Claude Opus 92\%, Gemini~3.8 Flash 90\%), yet fall well short on
the play tasks built from the same rules: knowing what each card does is not knowing which play
wins.

\emph{Judging a game is not computing a fight.} On the turn plans scored by playing the game
out, \agentname's policy answers 75\% in 0.03~s; the best LLM reaches 54\% with seconds of
reasoning, the others 19--43\%.
Its policy also leads on the choices cut from natural games (81\%).

\section{Lessons from Self-Improvement}
\label{sec:lessons}

The reality loop and the gates caught errors that no simulated metric revealed. We report six,
each found and corrected inside the loop; four of them also changed the research process.
\emph{(i) Copied games leak the future.} A search over game copies appeared to add 23.6 points of
first-place rate; the copies carried the random-number state, so the search saw the real future
shop and combat outcomes. Reseeded per sample, the gain vanished ($-0.65\pm1.14$ points).
\emph{(ii) Two errors can cancel.} The exam once matched live results only because its late-game
opponents were too strong while the agent's growth was overstated; once both were fixed, the exam
first-place rate rose to 43--49\%.
\emph{(iii) Winner's curse.} The best of many snapshots overstated its gain (+1.2--2.1 points on
selection seeds, $-$0.7 on unseen ones); hence gate \textsc{G4}.
\emph{(iv) Calibrate to the opponents you meet.} Measure a library on the opponents the learner
meets, not on uniform draws.
\emph{(v) The ladder catches what simulation praises.} TA-10 nearly doubled its predecessor's
exam first-place rate (24.2\% against 12.8\%), yet on the ladder the rating fell
from 8,968 to 7,658 within 32 games, 18 of which also had client failures (\cref{fig:ladder}). The
loop paused live testing, traced the fall to client failures and an early- and mid-game
weakness the exam had hidden, and retrained; TA-11 resumed the climb.
\emph{(vi) Two encoders drift apart.} Replays of live games showed the agent selling its whole
warband on one turn far more often than in simulation; one input flag was encoded differently by
the controller and the simulator after a trinket choice. The fix added a bit-level comparison of
both encoders to gate \textsc{G2}.

\section{Conclusion}
\label{sec:conclusion}

We presented \frameworkname, a self-improving research system: LLM research agents improve a
model, its training world and their own research process from the feedback of a real environment,
in a closed loop of training, research and live games. In 20 days it built \envname, a faithful,
full-content reproduction of Hearthstone Battlegrounds, and \agentname, which plays complete games
on the live ladder at the level of the server's best players. Asked to make the same decisions
directly, frontier LLMs, including the models that built the system, fall well short of it and
decide far too slowly for a live turn: LLM agents build far better than they play. We will
open-source \benchname and \envname at \mbox{\url{https://github.com/MCG-NJU/AutoTavern}};
limitations and ethics are discussed in \cref{app:deploy}.

{
\small
\bibliographystyle{plainnat}
\bibliography{references}
}

\appendix

\section{The \frameworkname system in detail}
\label{app:framework}
\paragraph{Conventions.} Days count from day 1 of the project; all work targets patch 36.6.3.

\paragraph{Agents.} The research ran as several long-lived agent sessions inside our agentic
research framework, each owning one role, plus short-lived sub-agents that a session dispatched
for a probe, a feature with its own test gate, or a check.

\begin{table}[h]
  \centering
  \caption{\textbf{Roles in \frameworkname.}}
  \label{tab:roles}
  \small
  \begin{tabular}{@{}lL{0.62\linewidth}@{}}
    \toprule
    Role & Responsibilities \\
    \midrule
    Research lead & reads the memory, chooses the next problem, dispatches work, records findings \\
    Environment & implements rules and content, verifies rules, fixes discrepancies found in live games \\
    Training & designs recipes and curricula, registers training runs and their successors \\
    Opponent modelling & builds and recalibrates the human-like opponent libraries \\
    Evaluation \& audit & evaluates within one build, re-checks candidates on unseen seeds, reviews \\
    Deployment & maintains the client controller and the models' input formats, collects live games \\
    Sub-agents & short tasks: probes, features behind a test gate, data analysis, reviews \\
    Human researcher & sets goals, rules on ambiguous mechanics, follows progress; in live games presses the shortcut key the AI names \\
    \bottomrule
  \end{tabular}
\end{table}

\paragraph{Shared research memory.} All coordination goes through versioned text files: a ledger of
numbered findings (date, configuration, readings, conclusion; corrections are new entries that
point back), a registry of 220 confirmed rule decisions and a list of open rule questions, a
hand-over note read at every change of hands, a persistent memory of lessons loaded at the start
of every session, and an experiment registry (about 10,000 experiments, each with its module, resources,
priority, dependencies and successors).

\paragraph{Fixes from live games.} Over the 20 days, discrepancies that live games exposed led to
184 fixes, each underlying defect counted once:
\begin{itemize}[leftmargin=*,itemsep=0pt,topsep=1pt]
  \item 35 of game rules and card behaviour in \envname, and 11 of card data and offer rules read
  from client records;
  \item 73 of the input encoding, where the controller and the simulator described the same state
  differently;
  \item 51 of the client controller: reading the game state, executing operations, recovery;
  \item 14 recalibrations of the opponent libraries and exams to the opponents met in live games.
\end{itemize}

\section{\envname details}
\label{app:env}

\begin{figure}[h]
  \centering
  \includegraphics[width=\linewidth]{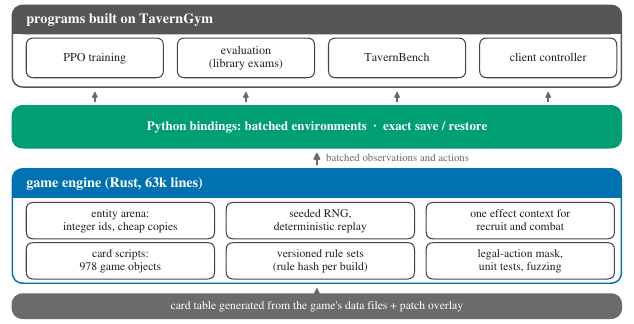}
  \caption{\textbf{\envname.} The Rust engine, its Python bindings and the programs built on them; the client controller feeds the live game to the agent in the same encoding.}
  \label{fig:engine}
\end{figure}

\paragraph{Architecture.} \envname is a Rust workspace of two crates: the game engine (about 63
thousand lines, of which 27 thousand are card scripts) and thin Python bindings whose
batched calls release the interpreter lock. Recruit phase and combat share one effect context; every
card is a table of event hooks; the state is an arena of entities with integer identifiers that is
cheap to copy and serialisable to JSON; all randomness comes from one seeded generator.
A source hash identifies every build and a rule hash every rule set, and a checkpoint trained
under another rule hash is not resumed.

\paragraph{Content.} The card table is generated from the game's data files; later
patches are applied as an overlay. All 978 objects of patch 36.6.3 (minions, heroes and their Hero Powers, spells, trinkets and
other cards) are implemented.

\paragraph{Verification.} Each mechanic was checked against the game data and official and
community sources and the check recorded (375 records); confirmed rulings go to the decision
registry (220) and unresolved ones to a list of open questions. Every mechanic has a unit test, and fuzzing plays random legal actions and checks state
invariants after every action.

\paragraph{RL interface.} 245 discrete actions (end turn, refresh, freeze, upgrade, Hero Power,
buy, sell, play or magnetize a hand card to a position, move a minion, choose among offers), each
with an exact legality mask. The observation contains only what the player sees, with opponents'
warbands as last seen in combat. A batched environment runs many eight-player games at once, and
running games can be saved and restored exactly.

\paragraph{Performance.} One core plays 190--280 complete eight-player games per second with scripted
bots and 52--76 thousand late-game combats per second.

\section{\agentname details}
\label{app:agent}

\begin{table}[h]
  \centering
  \caption{\textbf{\agentname releases.} TA-1 to TA-23 in release order, by day of the project, with
  the change each release introduced; the paper uses these names throughout.}
  \label{tab:releases}
  \footnotesize
  \begin{tabular}{@{}llL{0.25\linewidth}llL{0.25\linewidth}@{}}
    \toprule
    Release & Day & Change & Release & Day & Change \\
    \midrule
    TA-1 & 4 & first model & TA-13 & 13 & main line \\
    TA-2 & 5 & main line & TA-14 & 14 & opponents recalibrated to live games \\
    TA-3 & 6 & opponents fitted to human statistics; first-place goal & TA-15 & 14 & play-order fix \\
    TA-4 & 6 & main line & TA-16 & 14 & new opponent libraries \\
    TA-5 & 7 & main line & TA-17 & 15 & graded opponent libraries \\
    TA-6 & 7 & main line & TA-18 & 16 & environment update \\
    TA-7 & 8 & main line & TA-19 & 17 & recalibrated opponents \\
    TA-8 & 8 & main line & TA-20 & 17 & main line \\
    TA-9 & 9 & environment update & TA-21 & 17 & environment update \\
    TA-10 & 9 & opponents from recorded games & TA-22 & 18 & main line \\
    TA-11 & 10 & merged specialist models & TA-23 & 18 & final \\
    TA-12 & 13 & main line &  & &  \\
    \bottomrule
  \end{tabular}
\end{table}

\begin{figure}[h]
  \centering
  \includegraphics[width=\linewidth]{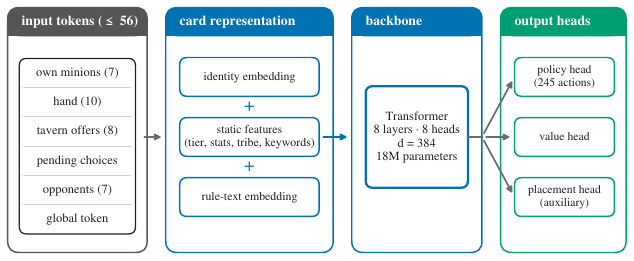}
  \caption{\textbf{\agentname's network.} Entity tokens, a card representation that includes the card's rule text, a transformer trunk and three heads.}
  \label{fig:model}
\end{figure}

\paragraph{Architecture.} Each own minion, hand card, tavern offer, pending choice and opponent is a
token (at most 56, plus one global token). A card is the sum of an identity embedding, a projection of
its static features (type, tier, stats, tribe, keywords, mechanics) and a projection of an embedding
of its rule text, which gives rare cards a sensible start. Attachments (magnetic minions, Dark Gifts)
are added to their minion's token, and the game-wide state (trinkets, persistent effects and
counters) is pooled into the global token. An opponent token holds the public leaderboard
information, the pooled warband last seen in combat and the result of that combat. Inputs are
compressed with a symmetric logarithm. A pointer head scores the 245 masked actions; a value head and
an auxiliary placement head share the trunk (d=384, 8 layers, 8 heads, 18M parameters).

\begin{table}[h]
  \centering
  \caption{\textbf{Training configuration of the final stage (TA-23).}}
  \label{tab:hparams}
  \small
  \resizebox{\linewidth}{!}{%
  \begin{tabular}{@{}ll@{\hspace{2em}}ll@{}}
    \toprule
    Games per member $\times$ steps & 512 $\times$ 256 & Iterations per stage & 300 \\
    Decisions per iteration & about $1.05\times10^6$ & Minibatch / epochs & 8,192 / 1 \\
    Learning rate & $6\times10^{-6} \to 5\times10^{-6}$ & Clip / target KL & 0.2 / 0.05 \\
    $\gamma$ / $\lambda$ & 1.0 / 0.995 & Gradient norm clip & 1.0 \\
    Entropy / value / auxiliary coefficients & 0.002 / 0.5 / 0.1 & KL anchor to the stage's start & 0.1 for 100 iterations \\
    \bottomrule
  \end{tabular}}
\end{table}

\paragraph{Model size.} An early experiment on day 3 trained three sizes of the network from
scratch with the same recipe and the same number of decisions (400 iterations of about 4.2 million
decisions each), measured as in \cref{tab:agent} (\cref{tab:model-size}). The largest is the best
on placement and top-four finishes in both exams, and \agentname uses it. First places are rare
in this table because these short runs from scratch predate the opponents and the environment of
the delivered versions; the comparison is between the sizes only.

\begin{table}[h]
  \centering
  \caption{\textbf{Model size} (an early experiment: from scratch, same recipe, 400 iterations
  each).}
  \label{tab:model-size}
  \footnotesize
  \setlength{\tabcolsep}{4pt}
  \begin{tabular}{@{}lcccccc@{}}
    \toprule
    & \multicolumn{3}{c}{Main exam} & \multicolumn{3}{c}{Harder exam} \\
    \cmidrule(lr){2-4}\cmidrule(l){5-7}
    Network & Place $\downarrow$ & First \% & Top-4 \% & Place $\downarrow$ & First \% & Top-4 \% \\
    \midrule
    3.5M parameters (d=192, 4 layers) & 3.84 & 0.3 & 73.9 & 4.47 & 0.3 & 54.0 \\
    8M parameters (d=256, 6 layers) & 3.77 & 0.5 & 75.6 & 4.40 & 0.3 & 56.2 \\
    18M parameters (d=384, 8 layers) & \textbf{3.67} & 0.5 & \textbf{78.5} & \textbf{4.28} & 0.4 & \textbf{59.6} \\
    \bottomrule
  \end{tabular}
\end{table}

\paragraph{Opponent libraries.} A library entry is one seat's board at one turn, with its hand and
hidden counters. Libraries were first built from statistics of top
human players (tribe shares, upgrade curves, purchase rates, warband strength by turn) and recorded
games; each turn a library seat takes an entry that matches a target strength curve. They were then
recalibrated to the opponents the agent actually met on the client: strength is measured on the
opponents the learner meets, not on entries drawn uniformly, and a graded set of libraries runs from
aligned ones to harder ones whose late boards carry venomous minions, hands and ceiling builds. Exam
libraries are never used for training; training uses twins built
the same way from other sources.

\paragraph{Training stages.} Learning started from scratch on a simplified rule set, moved to the full
content, then to the player's view, then to a first-place goal against opponents fitted to human
statistics, and then to library opponents. Later versions are chains of 300-iteration stages (about 310
million decisions and 570 thousand games each), each starting from the previous stage under a KL
anchor for its first 100 iterations. Later stages also used a graded ladder of training libraries; a version is delivered only after
a re-check on unseen seeds.

\begin{figure}[h]
  \centering
  \includegraphics[width=0.75\linewidth]{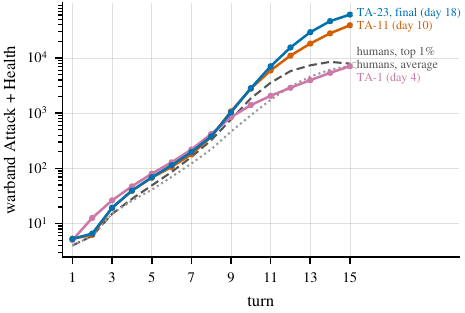}
  \caption{\textbf{Warband strength by turn.} \agentname versions on the main exam (three seeds) and
  human players of the same patch (public statistics), log scale.}
  \label{fig:turn-strength}
\end{figure}

\paragraph{Warband strength.} \Cref{fig:turn-strength} follows the total Attack and Health of the
warband turn by turn. The first version keeps pace with the average human player; the final one
matches the top 1\% of human players up to turn 9 and builds two to eight times their totals from
turn 11 on. Both sides are survivor curves, and the agent's opponents are library seats rather than
people, so the comparison describes growth, not head-to-head strength.

\paragraph{Versions head to head.} \Cref{tab:version-league} seats eight delivered versions in the
same lobbies, without library seats: every game has one seat for each version. The order of the
table is the order of release; the final version wins a quarter of the games, twice its fair share.

\begin{table}[h]
  \centering
  \caption{\textbf{Eight versions in the same lobbies} (three seeds, 24,576 games per version; a fair
  share is a placement of 4.5, 12.5\% first and 50\% top-four).}
  \label{tab:version-league}
  \small
  \begin{tabular}{@{}llccc@{}}
    \toprule
    Version & Day & Avg.\ place & First (\%) & Top-4 (\%) \\
    \midrule
    TA-23 & 18 & 3.95 & 26.2 & 57.9 \\
    TA-21 & 17 & 4.06 & 23.5 & 56.7 \\
    TA-17 & 15 & 4.36 & 19.7 & 51.6 \\
    TA-14 & 14 & 4.40 & 14.9 & 52.4 \\
    TA-11 & 10 & 4.53 & 9.4 & 51.9 \\
    TA-3 & 6 & 4.76 & 4.4 & 46.6 \\
    TA-2 & 5 & 4.85 & 1.2 & 44.7 \\
    TA-1 & 4 & 5.10 & 0.6 & 38.2 \\
    \bottomrule
  \end{tabular}
\end{table}

\begin{figure}[h]
  \centering
  \includegraphics[width=0.62\linewidth]{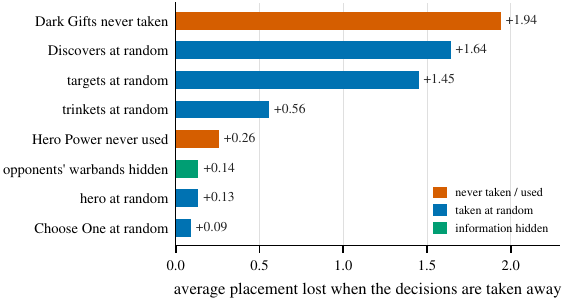}
  \caption{\textbf{Decision value.} Average placement lost by \agentname (TA-23) when one class of
  its decisions is taken away (made at random, never made, or the information hidden); main exam,
  three seeds, 24,576 games per bar.}
  \label{fig:decision-value}
\end{figure}

\paragraph{What the decisions are worth.} \cref{fig:decision-value} takes one class of the final
agent's decisions away at a time (made uniformly at random, or never taken) and measures the
placement lost against the main exam (three seeds, 24,576 games each). Dark Gifts, Discover
choices and targets are worth 1.4--1.9 places each, trinket choices half a place, and the Hero
Power a quarter.

\begin{figure}[h]
  \centering
  \includegraphics[width=0.75\linewidth]{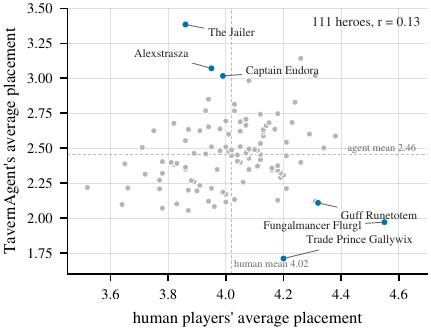}
  \caption{\textbf{Heroes.} Average placement of \agentname (TA-23) with each hero against human
  players' average placement with that hero (public statistics). Dashed lines:
  the two means; the six heroes where the two disagree most are named.}
  \label{fig:hero-table}
\end{figure}

\paragraph{Heroes.} Dealt every one of 111 heroes uniformly (three seeds, 24,576 games each),
\agentname's hero strengths differ from human players' (Pearson $r = 0.13$ between its placement
with a hero and theirs in public statistics, \cref{fig:hero-table}; absolute levels differ because
\agentname plays against library opponents): it is strongest with heroes people rate as weak
(Fungalmancer Flurgl, Trade Prince Gallywix, Guff Runetotem) and weakest, relative to people, with
heroes whose powers need planning across turns (The Jailer, Alexstrasza, Captain Eudora).

\section{Live deployment}
\label{app:deploy}

\paragraph{Setting and ethics.} All games were played in the solo mode of the live game on the
authors' own account, without commercial purpose, on days 5--18 of the project. The controller
feeds the live game's state to the current \agentname version in the environment's own encoding;
the AI decides every operation and names it, and a human presses the shortcut key that performs
that purchase, sale, play, choice or target, exactly as named, making no decision of their own.
Shortcut keys and state-based hints are both features of the game's official plug-in. When a
proposal was illegal in the client (an encoding or controller error), the game was abandoned and
the error fixed before play resumed, so no human judgement enters the games counted. The study
covers one game and one patch, with calibrated library opponents in training.

\paragraph{Records.} Each ladder result (392 games, read from the game's history screens) is
linked to the controller's decision logs, which name the model in play. The rating chain is
continuous from 6,000 to 14,007; four results without a recorded placement count in it but not in
the placement statistics (388 valid placements). \cref{tab:ladder} uses only the games that are
linked with high confidence to a single version, and \cref{tab:ladder-all} adds the games with a
client failure.

\begin{table}[h]
  \centering
  \caption{\textbf{Live-ladder games by \agentname version} (games linked to a
  single version with high confidence, without games with a client failure; ratings at the first and
  last game of each group). All games, failures included, are in \cref{tab:ladder-all}.}
  \label{tab:ladder}
  \small
  \begin{tabular}{@{}llrcccc@{}}
    \toprule
    Versions & Days & Games & Avg.\ place & First (\%) & Top-4 (\%) & Rating \\
    \midrule
    TA-2 & 5--6 & 26 & 3.46 & 3.8 & 76.9 & 6,000 $\to$ 7,177 \\
    TA-5--6 & 7--8 & 10 & 3.30 & 20.0 & 80.0 & 7,430 $\to$ 8,225 \\
    TA-8--10 & 8--10 & 47 & 3.57 & 12.8 & 66.0 & 8,117 $\to$ 8,348 \\
    TA-11 & 11--13 & 87 & 3.10 & 16.1 & 78.2 & 8,348 $\to$ 10,910 \\
    TA-12--14 & 13--16 & 28 & 3.36 & 7.1 & 78.6 & 11,001 $\to$ 12,329 \\
    TA-16--17 & 15--16 & 30 & 3.67 & 16.7 & 70.0 & 11,995 $\to$ 12,727 \\
    TA-19--23 & 17--18 & 46 & \textbf{2.37} & \textbf{32.6} & \textbf{91.3} & 12,727 $\to$ \textbf{14,007} \\
    \bottomrule
  \end{tabular}
\end{table}

\paragraph{Comparison with the top players.} The rating measures accumulated wins as much as
strength: it rises only as fast as games are played, and the ladder was played for two weeks. We
therefore compare recent results on one measure for every player: the public leaderboard tracker
\citep{wallii} records each player's rating updates and estimates a placement for each from the
rating change (a record can cover one or several games; ties are estimated as $x.5$). The estimate is
the same for every player, so it compares players fairly even though it is not a count of games; we
do not mix it with our own logs, which the other players do not have. The agent met the same top
players: in the 20 games whose opponents could be looked up on the leaderboard,
the 29 opponents found were rated 12,190--17,583 (median 13,294).

The logged record ends at 14,007. The loop kept playing, and \cref{tab:top10} compares the last
20 leaderboard records of every player at the cut-off, when the agent was rated about 15,000 and in
the server's top 100. Its average placement, 1.60 with 80\% first places, is
better than that of each of the ten highest-rated players (1.88--3.03; 2.34 and 59\% together),
and its rating gain per record (+64) is nine times theirs (+7). On every statistic of this
window the agent is ahead of the ten highest-rated players, and at the cut-off its rating was
still rising steeply with no sign of a plateau: it plays at the level of the server's best.

\begin{table}[h]
  \centering
  \caption{\textbf{The agent against the server's ten highest-rated players}: each player's last 20
  records on the public leaderboard \citep{wallii} at the cut-off (placements estimated
  by the leaderboard from rating changes, the same measure for every player; players anonymised).}
  \label{tab:top10}
  \small
  \begin{tabular}{@{}lcccr@{}}
    \toprule
    Player & Avg.\ place & First (\%) & Top-4 (\%) & Rating change \\
    \midrule
    P1 & 3.03 & 50 & 65 & $-$319 \\
    P2 & 2.20 & 70 & 80 & +289 \\
    P3 & 2.43 & 55 & 85 & $-$88 \\
    P4 & 1.88 & 65 & 90 & +487 \\
    P5 & 2.50 & 55 & 75 & $-$76 \\
    P6 & 2.05 & 70 & 85 & +459 \\
    P7 & 2.20 & 60 & 80 & +67 \\
    P8 & 2.53 & 50 & 80 & +44 \\
    P9 & 2.53 & 45 & 85 & +140 \\
    P10 & 2.05 & 70 & 85 & +422 \\
    Top ten together (200 records) & 2.34 & 59 & 81 & +1,425 \\
    \midrule
    \agentname (rating $\approx$15,000) & \textbf{1.60} & \textbf{80} & \textbf{90} & \textbf{+1,282} \\
    \bottomrule
  \end{tabular}
\end{table}

\paragraph{Client failures.} The controller logs flag, per game, events of the client rather than the
agent: the client stopped, a recruit turn was left unfinished, retries ran out, or the client
reconnected during combat. 96 of the 388 games with a placement carry at least one; they are
identified from these flags alone, before looking at the result. \Cref{tab:ladder-all} gives the
version groups with these games included. Excluding them improves the overall and the latest
groups' results and worsens some earlier groups', since a reconnection is more likely in a long game.

\begin{table}[h]
  \centering
  \caption{\textbf{All live-ladder games by version}, client failures included.}
  \label{tab:ladder-all}
  \small
  \begin{tabular}{@{}lrcccr@{}}
    \toprule
    Versions & Games & Avg.\ place & First (\%) & Top-4 (\%) & With failure \\
    \midrule
    TA-2 & 40 & 3.38 & 2.5 & 77.5 & 14 \\
    TA-5--6 & 24 & 3.12 & 20.8 & 79.2 & 14 \\
    TA-8--10 & 68 & 4.22 & 8.8 & 54.4 & 21 \\
    TA-11 & 89 & 3.07 & 16.9 & 78.7 & 2 \\
    TA-12--14 & 38 & 3.08 & 5.3 & 84.2 & 10 \\
    TA-16--17 & 48 & 3.15 & 20.8 & 79.2 & 18 \\
    TA-19--23 & 56 & 2.54 & 30.4 & 87.5 & 10 \\
    \bottomrule
  \end{tabular}
\end{table}

\section{\benchname details}
\label{app:bench}

\begin{table}[h]
  \centering
  \caption{\textbf{\benchname categories.} 40 scenarios, 960 tasks. \agentname answers
  through its own interface (\cref{sec:bench}); the ten rule-execution scenarios have no agent
  counterpart, because \envname executes the rules for it.}
  \label{tab:bench-cats}
  \small
  \begin{tabular}{@{}lL{0.46\linewidth}L{0.13\linewidth}l@{}}
    \toprule
    Category & Scenarios & Label & \agentname \\
    \midrule
    Rules (7) & legal actions; one-on-one fights; battlecry, spell and magnetic results;
      end- and start-of-turn effects; triples & executed & one-on-one \\
    Economy (5) & gold accounting; upgrade price; can I triple now; gold chains; best purchase
      for the next combat & executed / simulated & purchase \\
    Combat (6) & who attacks first; deterministic fights; win / tie / loss; expected damage;
      lethal risk; start-of-combat effects & simulated & all \\
    Positioning (5) & best of five arrangements; taunt and shield; deathrattle and reborn; against
      venomous; against cleave & simulated & all \\
    Ordering (5) & sell timing; play triggers; spells and minions; adjacent battlecries; mixed & simulated & all \\
    Targets (5) & spell, activate, Hero Power and magnetic targets; which minion to sell & simulated & all \\
    Selection (5) & hero; trinket; discover; choices in natural games; which card to play &
      human, simulated, play-out & all \\
    Strategy (2) & turn plans, early game (turns 4--7) and late game (turns 8+): upgrade, buy,
      refresh or save the Gold & play-out & all \\
    \bottomrule
  \end{tabular}
\end{table}

\paragraph{Scenes.} Every static task starts from a real game state that the environment reshapes for its scenario
(turn, tier, gold, health, and warbands drawn from the current card pool by the scenario's
filter), with heroes whose Hero Powers have no effect in the scene, so that no hidden effect
changes a label. Strategy and natural-choice tasks are instead
cut from games of the main exam lobby (\cref{tab:agent}) at decisions of the kind.

\paragraph{Labels and scoring.} Rule tasks execute the stated actions on copies of the scene. A task
whose outcome involves random numbers is kept only when two or three reseeded copies agree.

Combat tasks use 4,096 simulated combats (the damage-distribution scenario 400 single combats), and
arrangement tasks 2,048 combats per arrangement. A probability triple is correct when its most likely
outcome matches and its Brier score is at most 0.1 (graded $1-\text{Brier}/2$).

Decision tasks label every option by 1,024 combats after it. A task is kept when the choice matters:
the best option beats the median one by 5--8 points of win rate, and at most a quarter of the
options are within three points of the best. An answer is correct when its option is within three
points of the best, and its graded score is its win rate over the best one's.

Strategy tasks are cut at the first decision of a turn in natural games of the main-exam lobby. Four
plans (upgrade the Tavern; buy the minion the labeller prefers; refresh first; end the turn and keep
the Gold), each legal in the state, are each played out 128 times by the labeller (TA-21), which
plays the rest of the turn and the game as it sees fit. A task is kept when the best plan's mean
placement beats the second-best's by 0.25 places and two standard errors of the difference, and the
letters A--D rotate so that each is the answer about equally often (chance 25\%). Ending the turn is
never the best plan; the contest is among the other three. The labels are outcomes of play, not the
labelling agent's preferences: answering with its own policy, the labeller (TA-21) reaches 73\% on
these tasks, the evaluated TA-23 75\%, and the first version TA-1 57\%.

Hero and trinket tasks use the average placement of top human players in public statistics and keep
offers whose best option leads by 0.1 (heroes) or 0.15 (trinkets) places.

\paragraph{Prompts.} One system prompt for all tasks; the user prompt has a rules summary (economy,
triples, combat order, keywords, damage), the player-visible state, the rule text of every card
shown, the question with its objective and the JSON answer format. A target task, abridged:

\begin{quote}\small\ttfamily\frenchspacing\raggedright
Turn 6. Your health 30. Tavern tier 5. Gold 10.\\
Your warband: 0. Bronze Timewalker (Dragon, tier 4) 4/5; 1. Twilight Tidehunter (Murloc, tier
4) 4/6; \dots\\
Opponent warband: 0. Very Hungry Winterfinner (Murloc, tier 2) 2/6 [Taunt]; 1. Greedy Conniver
(Pirate, tier 3) 7/7; \dots\\
Card rules: \dots\ Boon of Light: Give a minion Divine Shield.\\
Your hero is George the Fallen; you use the Hero Power on one minion of your warband: which one
(warband position)? Your next combat is against the opponent warband shown.\\
Answer format: \{"answer": <warband position>\}
\end{quote}

\begin{table}[h]
  \centering
  \caption{\textbf{Example tasks}, one per category (questions abridged; each task also shows the
  state and the rule text of every card). Labels come from \envname.}
  \label{tab:bench-examples}
  \footnotesize
  \begin{tabular}{@{}lL{0.55\linewidth}L{0.27\linewidth}@{}}
    \toprule
    Category & Question & Label \\
    \midrule
    Rules & You cast Misplaced Tea Set. List the Attack and Health of every minion afterwards. &
      [[9,6],[13,8]] (executed, three reseeded copies agree) \\
    Economy & Which tavern minions do you buy, in order, with 9 Gold, each played at once, to win
      the next combat against the warband shown? & positions 2,1,4 (win rate 8.9\%, best of 157
      sequences) \\
    Combat & Estimate the probabilities that you win, tie or lose this combat. &
      0.75 / 0.20 / 0.04 (4,096 combats) \\
    Positioning & Which of five arrangements of your warband (against a Cleave opponent) wins most
      often? & E (best by $\geq$5 points over 2,048 combats each) \\
    Ordering & In which order do you play three minions and sell one before the next combat? &
      B, D, C, A (win rate 90\%) \\
    Targets & Onto which of your Mechs do you magnetize Enchanted Sentinel? & position 0 (71\%) \\
    Selection & Which of three Discovered minions do you take and play? & Electric Synthesizer
      (19\%) \\
    Strategy & Turn 4, 6 Gold, upgrade price 5: upgrade, buy Strike Oil, refresh, or keep the
      Gold? & buy (placement 1.48 against 1.88 for the next best; 128 play-outs per plan) \\
    \bottomrule
  \end{tabular}
\end{table}

\paragraph{Language models.} All models were queried through their APIs with the benchmark's
system prompt and no tools, streamed, within the output budget; where an API does not stop
generation at the budget, answers over it are scored as cut off.

\paragraph{Modes and budget.} A pilot with one task from each scenario chose each model's mode: Kimi-K2.6, DeepSeek-V4.1-Flash and
Intern-S2-Preview answer with reasoning off, the other models with low-effort reasoning. In the
evaluation
48\% of Grok~4.7's, 23\% of Kimi-K2.6's and 16\% of Gemini~3.8 Flash's answers were over the
budget, at most 4\% of the other models'. Per task the LLMs used 0.9--3.9k input tokens and 9--2,772 output tokens (Grok~4.7: 6,625).

\paragraph{Latency.} Measured wall-clock times include queueing and start-up and depend on the
access path, so the paper times every LLM on one standard footing (\cref{sec:bench}). A task's time
is the median time to the first answer token of the model's first-party API, in the variant that
matches our setting, plus the task's answer at the median output speed, both as published by
Artificial Analysis \citep{artificialanalysis}. The first-answer time includes the model's typical
thinking; a model without a listing takes the medians of the listed ones. On our tasks some models
think far longer than that: Grok~4.7 generates 3,600 tokens for the median task even at low
effort. Timed with their own output tokens instead, the median task takes 3--50~s, and every
real-time score stays the same.

\paragraph{Decision model.} Jev \citep{typesafejev} takes a state and typed questions and returns, per
question, the chosen option with a probability for every option; it does not write free text. Its
state is the prompt's rules summary, player view and card rules, its instructions the task's
question, and every task is posed as a choice over candidates:
\begin{itemize}[leftmargin=*,itemsep=0pt,topsep=1pt]
  \item choice tasks: the options the task lists; target tasks: every warband or hand position;
  \item ordering tasks: every order of the listed actions (up to 24); purchase tasks: every purchase
  sequence the Gold allows (up to 157);
  \item probability tasks: win, tie and loss, or die and survive, whose probabilities are the answer;
  \item numeric tasks: the integers 0--30 for Gold and damage (the expected damage is the mean of its
  distribution) and 0--60 for each Attack and Health; legality tasks: legal or illegal for each action.
\end{itemize}
The candidate lists hold every option and none of the labels' values, and the common scorer applies.
Jev answered every task without an error in 0.26~s per task (median, one request per task) with
1.9k input tokens; its billed output tokens count option probabilities rather than generated text,
so the output budget does not apply.

\paragraph{Full set.} On all 960 tasks \agentname answers 92\% of the play tasks, against 91\% on
the 368 tasks that every system answers.

\paragraph{LLMs with \envname as a tool.} To separate what the simulator contributes from what
the policy contributes, five LLMs answered the short-term decision tasks (positioning, ordering,
targets, purchases, discover and hand choices) with two tools over the task's state in \envname:
\texttt{test\_play(actions)} applies recruit actions in the engine and fights the next combat 256
times, \agentname's budget per candidate, and \texttt{inspect(actions)} shows the resulting
state; at most twelve calls per task, the same prompt otherwise. \Cref{tab:tools} compares each
model with its own text-only answers on the same tasks. The tools close most of the gap to
\agentname, and the remainder is the models' own: illegal or mis-indexed actions (hand positions
shift after a play), candidates left untested within the budget, and purchase sequences, where
the candidate space is largest. Every model answered all 144 tasks (eight per scenario); Kimi-K2.6,
Intern-S2-Preview and DeepSeek-V4.1-Flash ran with reasoning off, the others at low effort.

\begin{table}[h]
  \centering
  \caption{\textbf{LLMs with \envname as a tool} (accuracy in \%, text-only / with tools per
  category; the agent column is \agentname on the same tasks).}
  \label{tab:tools}
  \footnotesize
  \setlength{\tabcolsep}{3pt}
  \begin{tabular}{@{}lrrrrrrrrrrr@{}}
    \toprule
    Model & Tasks & Text & Tools & Position & Order & Targets & Purchase & Select & Agent & Calls & Time (s) \\
    \midrule
    GPT-6 Astra & 144 & 62 & 97 & 55/98 & 72/100 & 55/92 & 75/88 & 69/100 & 99 & 5.2 & 21 \\
    GLM-5.3 & 144 & 38 & 81 & 20/82 & 40/62 & 38/95 & 38/62 & 75/94 & 99 & 5.8 & 11 \\
    Kimi-K2.6 & 144 & 26 & 65 & 8/68 & 22/35 & 38/98 & 25/0 & 50/88 & 99 & 8.9 & 39 \\
    Intern-S2-Preview & 144 & 32 & 41 & 12/40 & 32/30 & 40/45 & 12/12 & 69/75 & 99 & 9.8 & 18 \\
    DeepSeek-V4.1-Flash & 144 & 24 & 64 & 18/50 & 15/42 & 32/92 & 38/25 & 31/100 & 99 & 8.6 & 42 \\
    \bottomrule
  \end{tabular}
\end{table}

\section{Compute resources}
\label{app:compute}
Main-line training ran on eight H200 GPUs with 128 CPU cores. Training the 23 delivered
\agentname stages took about 1,900 GPU-hours; all other training runs of the 20 days
(explorations, ablations and abandoned lines) about 16,700 GPU-hours, and evaluations and analyses
about 1,300 GPU-hours (from the experiment registry over the 20 days, every run counted at its
configured size). The research agents ran on frontier LLMs, the GPT models GPT-6.1 Sol and
GPT-6 Astra and Claude Opus 5.5, inside our agentic research framework, and together they
generated about 100 million output tokens over the 20 days.

\end{document}